\documentclass[10pt,twocolumn]{article}

\PassOptionsToPackage{numbers, compress}{natbib}

\usepackage{cvpr}
\usepackage{times}
\usepackage{epsfig}
\usepackage{graphicx}
\usepackage[utf8x]{inputenc} 
\usepackage[T1]{fontenc}    
\usepackage{url}            
\usepackage{booktabs}       
\usepackage{amsfonts}       
\usepackage{nicefrac}       
\usepackage{microtype}      
\usepackage[ruled,vlined]{algorithm2e}
\usepackage{graphicx}
\usepackage{subcaption}
\usepackage{amsmath}
\usepackage{amsthm}
\usepackage{amssymb}
\usepackage{mathrsfs}
\usepackage{bm}
\usepackage[ruled,vlined]{algorithm2e}
\usepackage{multirow}
\usepackage{multicol}
\usepackage{slashbox}

\theoremstyle{plain}

\usepackage{breqn}

\usepackage[pagebackref=true,breaklinks=true,letterpaper=true,colorlinks,bookmarks=false]{hyperref}

\newcommand*{\affaddr}[1]{#1} 
\newcommand*{\affmark}[1][*]{\textsuperscript{#1}}
\newcommand*{\email}[1]{\texttt{#1}}

\cvprfinalcopy 

\def\cvprPaperID{2330} 

\begin{document}

\title{Deep Regionlets for Object Detection}


\author{%
Hongyu Xu\affmark[1]\thanks{Work started during an internship at Snap Research}~,~Xutao Lv\affmark[2],~Xiaoyu Wang\affmark[2],~Zhou Ren\affmark[3],~Navaneeth Bodla\affmark[1] and Rama Chellappa\affmark[1]\\
\affaddr{\affmark[1]Department of Electrical and Computer Engineering and the Center for Automation Research, UMIACS \\
University of Maryland, College Park, MD, USA}\\
\affaddr{\affmark[2]Intellifusion~~~~~~\affmark[3]Snap Inc.}\\
\email{\affmark[1]hyxu@umiacs.umd.edu \affmark[2]lvxutao@gmail.com \affmark[2]fanghuaxue@gmail.com} \\
\email{\affmark[3]zhou.ren@snap.com \affmark[1]nbodla@umiacs.umd.edu \affmark[1]rama@umiacs.umd.edu} \\
}

\maketitle

\begin{abstract}

In this paper, we propose a novel object detection framework named "Deep Regionlets" by establishing a bridge between deep neural networks and conventional detection schema for accurate generic object detection. Motivated by the abilities of regionlets for modeling object deformation and multiple aspect ratios, we incorporate regionlets into an end-to-end trainable deep learning framework. The deep regionlets framework consists of a region selection network and a deep regionlet learning module. Specifically, given a detection bounding box proposal, the region selection network provides guidance on where to select regions to learn the features from. The regionlet learning module focuses on local feature selection and transformation to alleviate local variations. To this end, we \emph{first} realize \emph{non-rectangular} region selection within the detection framework to accommodate variations in object appearance. Moreover, we design a ``gating network" within the regionlet leaning module to enable soft regionlet selection and pooling. The Deep Regionlets framework is trained end-to-end without additional efforts. We perform ablation studies and conduct extensive experiments on the PASCAL VOC and Microsoft COCO datasets. The proposed framework outperforms state-of-the-art algorithms, such as RetinaNet and Mask R-CNN, even without additional segmentation labels. 

\end{abstract}

\section{Introduction}

Generic object detection has been extensively studied by the computer vision community over several decades~\cite{Huang_HRSZKFFWSGM_CVPR2017,Bodla_BSCD_ICCV2017,Wang_WYZL_2015TPAMI,Girshick_GDDM_2014CVPR,Girshick_G_ICCV2015,Ren_RHGS_TPAMI2016,Dai_DLHS_NIPS2016,Yu_YCMD_BMVC2016,Lin_LGGHD_ICCV2017,Wang_WYZL_ICCV2013,Viola_VJ_CVPR2001,Dalal_DT_CVPR2005,Felzenszwalb_FGMR_TPAMI2010,Cai_CV_CVPR2018,Singh_BD_CVPR2018,Zhang_ZWBLL_CVPR2018} due to its appeal to both academic research explorations as well as commercial applications. Given an image of interest, the goal of object detection is to predict the locations of objects and classify them at the same time. The key challenge of the object detection task is to handle variations in object scale, pose, viewpoint and even part deformations when generating the bounding boxes for specific object categories. 

Numerous methods have been proposed based on hand-crafted features (\textit{i.e.} HOG~\cite{Dalal_DT_CVPR2005}, LBP~\cite{Ahonen_AHP_ECCV2004}, SIFT~\cite{Lowe_L_ICCV1999}). These approaches usually involve an exhaustive search for possible locations, scales and aspect ratios of the object, by using the sliding window approach. However, Wang \textit{et al}.'s~\cite{Wang_WYZL_ICCV2013} regionlet-based detection framework has gained a lot of attention as it provides the flexibility to deal with different scales and aspect ratios without performing an exhaustive search. It first introduced the concept of \textbf{regionlet} by defining a three-level structural relationship: candidate bounding boxes (sliding windows), regions inside the bounding box and groups of regionlets (sub-regions inside each region). It operates by directly extracting features from regionlets in several selected regions within an arbitrary detection bounding box and performs (max) pooling among the regionlets. Such a feature extraction hierarchy is capable of dealing with variable aspect ratios and  flexible feature sets, which leads to improved learning of robust feature representation of the object for region-based object detection. 

\begin{figure*}
\centering{
\includegraphics[width=1.88\columnwidth]{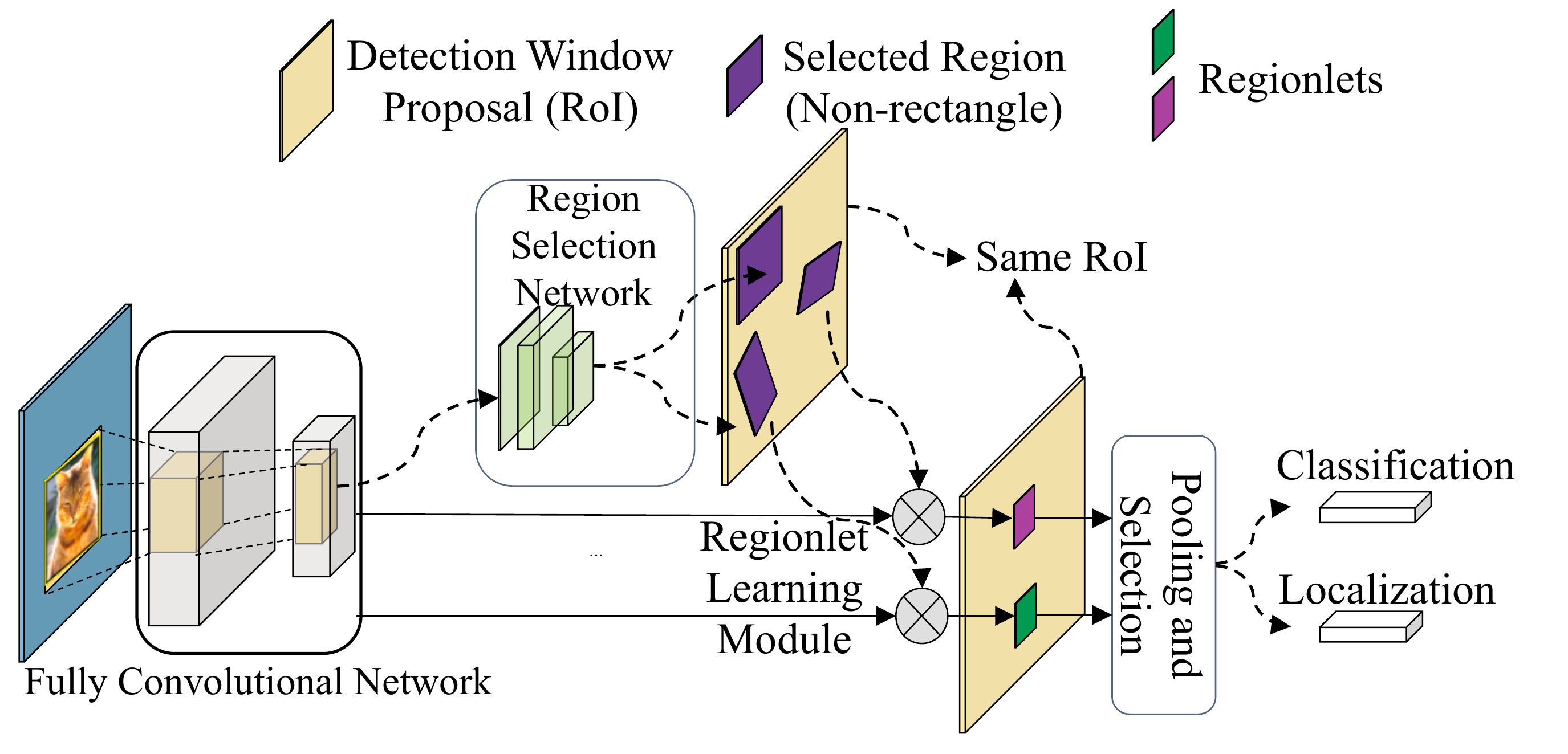}}
\caption{Architecture of the Deep Regionlets detection framework. It consists of a region selection network (RSN) and a deep regionlet learning module. The region selection network performs \emph{non-rectangular} region selection from the detection window proposal generated by the region proposal network. Deep regionlet learning module learns the regionlets through a spatial transformation and a gating network. The entire pipeline is end-to-end trainable. For better visualization, the region proposal network is not displayed here.}
\label{fig::DeepRegionlet_overall}
\end{figure*}

Recently, deep learning has achieved significant success on many computer vision tasks such as image classification~\cite{Krizhevsky_KSH_2012NIPS,He_HZRS_CVPR16,Ranjan_RBXSCCC_2018}, semantic segmentation~\cite{Long_LSD_CVPR2015} and object detection~\cite{Girshick_GDDM_2014CVPR} using the deep convolutional neural network (DCNN) architecture. Despite the excellent performance of deep learning-based detection framework, most network architectures~\cite{Ren_RHGS_TPAMI2016,Dai_DLHS_NIPS2016,Liu_LAESRFB_CVPR2016} do not take advantage of successful conventional ideas such as deformable part-based model (DPM) or \textit{regionlets}. Those methods have been effective for modeling object deformation, sub-categories and multiple aspect ratios. Recent advances~\cite{Ouyang_OZWQLTLYWL_TPAMI2017,Dai_DQXLZHW_ICCV2017,Mordan_MTCH_2017} have achieved promising results by combining the conventional DPM-based detection methodology with deep neural network architectures.

These observations motivate us to establish a bridge between deep convolutional neural network and conventional object detection schema. In this paper, we incorporate the conventional Regionlet method into an end-to-end trainable deep learning framework. Despite being able to handle arbitrary bounding boxes, several drawbacks arise when directly integrating the regionlet methodology into the deep learning framework. 
First, in ~\cite{Wang_WYZL_ICCV2013}, Wang \textit{et al}. proposed to learn cascade object classifiers after hand-crafted feature extraction in each regionlet. However, end-to-end learning is not feasible in this framework. Second, regions in regionlet-based detection have to be rectangular, which does not effectively model the deformations of an object which results in variable shapes. Moreover, both regions and regionlets are fixed after training is completed. 

To this end, we propose a novel object detection framework named "Deep Regionlets" to integrate the deep learning framework into the traditional regionlet method~\cite{Wang_WYZL_ICCV2013}. The overall design of the proposed detection system is illustrated in Figure~\ref{fig::DeepRegionlet_overall}. It consists of a region selection network (RSN) and a deep regionlet learning module. The region selection network performs \emph{non-rectangular} region selection from the detection window proposal\footnote{The detection window proposal is generated by a region proposal network (RPN)~\cite{Ren_RHGS_TPAMI2016,Dai_DLHS_NIPS2016,Girshick_G_ICCV2015}. It is also called region of interest (ROI)} (RoI) to address the limitations of the traditional regionlet approach.
We further design a deep regionlet learning module to learn the regionlets through a spatial transformation and a gating network. By using the proposed gating network, which is a soft regionlet selector, the resulting feature representation is more effective for detection. The entire pipeline is end-to-end trainable using only the input images and ground truth bounding boxes.

We conduct a detailed analysis of our approach to understand its merits and evaluate its performance. Extensive experiments on two detection benchmark datasets, PASCAL VOC~\cite{Everingham_EGWWZ_IJCV2010} and Microsoft COCO~\cite{Lin_LMBHPRDZ_ECCV2014} show that the proposed deep regionlet approach outperforms several competitors~\cite{Ren_RHGS_TPAMI2016,Dai_DLHS_NIPS2016,Dai_DQXLZHW_ICCV2017,Mordan_MTCH_2017}. Even without segmentation labels, we outperform state-of-the-art algorithms such as Mask R-CNN~\cite{He_HGDG_ICCV2017} and RetinaNet~\cite{Lin_LGGHD_ICCV2017}. To summarize, we make the following contributions: 

\begin{itemize}
\item We propose a novel deep regionlet approach for object detection. Our work extends the traditional regionlet method to the deep learning framework. The system is trainable in an end-to-end manner.
\item We design the RSN, which \textbf{first} performs \textbf{non-rectangular} region selection within the detection bounding box generated from a detection window proposal. It provides more flexibility in modeling objects with variable shapes and deformable parts.
\item We propose a deep regionlet learning module, including feature transformation and a gating network. The gating network serves as a soft regionlet selector and lets the network focus on features that benefit detection performance.
\item We present empirical results on object detection benchmark datasets, demonstrating superior performance over state-of-the-art.
\end{itemize}

\section{Related Work}

Many approaches have been proposed for object detection including both traditional ones~\cite{Felzenszwalb_FGMR_TPAMI2010,Wang_WYZL_ICCV2013,Viola_VJ_CVPR2001} and deep learning-based approaches~\cite{Girshick_G_ICCV2015,Ren_RHGS_TPAMI2016,Liu_LAESRFB_CVPR2016,Redmon_RDGF_CVPR2016,Dai_DLHS_NIPS2016,Girshick_GDDM_2014CVPR,He_HZRS_ECCV20014,Dai_DQXLZHW_ICCV2017,Mordan_MTCH_2017,Cai_CV_CVPR2018,Hu_HGZDW_CVPR2018,Zhao_ZLW_CVPR2018,Zhou_ZNGHX_CVPR2018,Zhang_ZQXSWY_CVPR2018,Zhang_ZWBLL_CVPR2018,Wang__WWGLZ_CVPR2018,Singh_BD_CVPR2018}. Traditional approaches mainly used hand-crafted features to train the object detectors using the sliding window paradigm. One of the earliest works~\cite{Viola_VJ_CVPR2001} used boosted cascaded detectors for face detection, which led to its wide adoption. Deformable Part Model-based detection (DPM)~\cite{Felzenszwalb_FGM_CVPR2010} proposed the concept of deformable part models to handle object deformations. Due to the rapid development of deep learning techniques~\cite{Krizhevsky_KSH_2012NIPS,He_HZRS_CVPR16,Simonyan_SZ_2014,Bodla_BZXCCC_WACV2017,Zhang_ZZLSWL_ICCV2017,Ranjan_RBXSCCC_2018,Xu_XZAC_2018,Bansal_BSSCD_2018,Wu_WBSNCD_2018}, the deep learning-based detectors have become dominant object detectors. 

Deep learning-based detectors could be further categorized into single-stage detectors and two-stage detectors, based on whether the detectors have proposal-driven mechanism or not. The single-stage detectors~\cite{Sermanet_SEZMFL_ICLR2014,Redmon_RDGF_CVPR2016,Liu_LAESRFB_CVPR2016,Fu_FLRTB_2017,Lin_LDGHHB_CVPR2017,Lin_LGGHD_ICCV2017,Zhang_ZWBLL_CVPR2018,Zhang_ZQXSWY_CVPR2018,Law_LD_2018} apply regular, dense sampling windows over object locations, scales and aspect ratios. By exploiting multiple layers within a deep CNN network directly, the single-stage detectors achieved high speed but their accuracy is typically low compared to two-stage detectors. 

Two-stage detectors~\cite{Girshick_G_ICCV2015,Ren_RHGS_TPAMI2016,Dai_DLHS_NIPS2016} involve two steps. They first generate a sparse set of candidate proposals of detection bounding boxes by the Region Proposal Network (RPN). After filtering out the majority of negative background boxes by RPN, the second stage classifies the proposals of detection bounding boxes and performs the bounding box regression to predict object categories and their corresponding locations. The two-stage detectors consistently achieve higher accuracy than single-stage detectors and numerous extensions have been proposed~\cite{Dai_DQXLZHW_ICCV2017,Mordan_MTCH_2017,He_HGDG_ICCV2017,Cai_CV_CVPR2018,Singh_BD_CVPR2018,Hu_HGZDW_CVPR2018,Cheng_CWSFXH_2018}. Our method follows the two-stage detector architecture by taking advantage of RPN without requiring dense sampling of object locations, scales and aspect ratios.  

\section{Our Approach}

In this section, we first review the traditional regionlet-based detection methods and then present the overall design of the end-to-end trainable deep regionlet approach. Finally, we discuss in detail each module in the proposed end-to-end deep regionlet approach. 

\subsection{Traditional Regionlet-based Approach}

\begin{figure}[!tbh]
\centering{
\includegraphics[width=0.63\columnwidth]{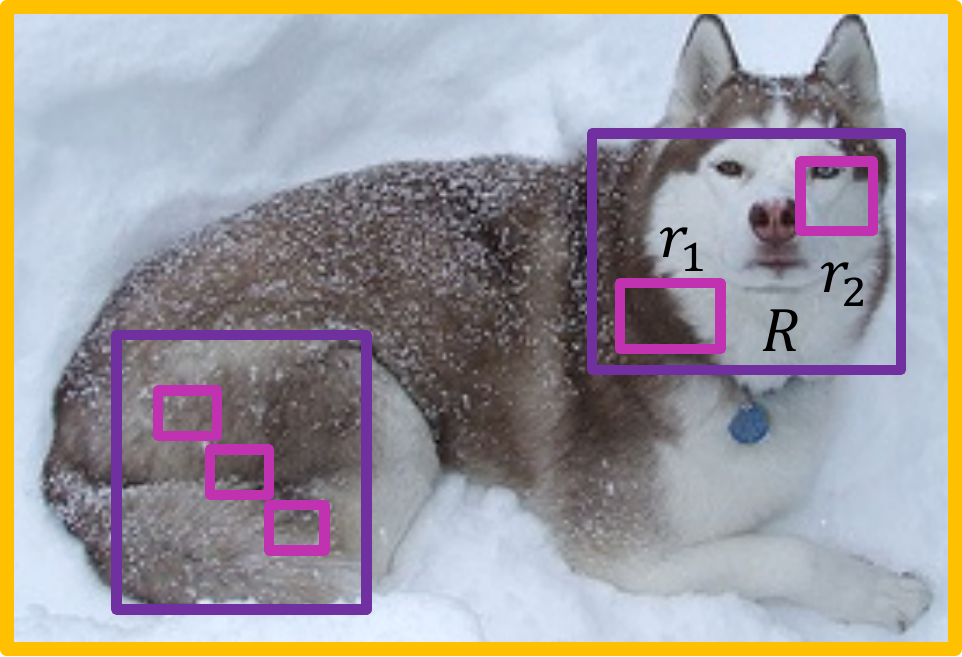}}
\caption{Illustration of structural relationships among the detection bounding box, feature extraction regions and regionlets. The yellow box is a detection bounding box and $R$ is a feature extraction region shown as a purple rectangle with filled dots inside the bounding box. Inside $R$, two small sub-regions denoted as $r_1$ and $r_2$ are the \emph{regionlets}.} 
\label{fig::regionlet_illustration}
\end{figure}

A \emph{regionlet} is a base feature extraction region defined proportionally to a window (\textit{i.e.} a sliding window or a detection bounding box) at arbitrary resolution (\textit{i.e.} size and aspect ratio). Wang \textit{et al.}~\cite{Wang_WYZL_ICCV2013} first introduced the concept of regionlet, as illustrated in Figure~\ref{fig::regionlet_illustration}. It defines a three-level structure among a detecting bounding box, number of regions inside the bounding box and a group of regionlets (sub-regions inside each region). In Figure~\ref{fig::regionlet_illustration}, the yellow box is a detection bounding box.  
$R$ is a rectangular feature extraction region inside the bounding box. Furthermore, small sub-regions $r_{i \{i = 1\dots N\}}$(\textit{e.g.} $r_1$, $r_2$) are chosen within region $R$, where we define them as a set of \emph{regionlets}. 

The difficulty of the arbitrary detection bounding box has been well addressed by using the \emph{relative} positions and sizes of regionlets and regions. However, in the traditional approach, the initialization of regionlets possess randomness and both regions ($R$) and regionlets (\textit{i.e.} $r_1$, $r_2$) are fixed after the training. Moreover, it is based on hand-crafted features (\textit{i.e.} HOG~\cite{Dalal_DT_CVPR2005} or LBP~\cite{Ahonen_AHP_ECCV2004}) in each regionlet respectively and hence not end-to-end trainable. To this end, we propose the following deep regionlet-based approach to address such limitations.

\subsection{System Architecture}

Generally speaking, an object detection network performs a sequence of convolutional operations on an image of interest using a deep convolutional neural network. At some layer, the network bifurcates into two branches. One branch, RPN generates a set of candidate bounding boxes\footnote{\cite{Ren_RHGS_TPAMI2016,Dai_DLHS_NIPS2016,Girshick_G_ICCV2015} also called the detection bounding box as detection window proposal} while the other branch performs classification and regression by pooling the convolutional features inside the proposed bounding box generated by the region proposal network~\cite{Ren_RHGS_TPAMI2016,Dai_DLHS_NIPS2016}. Taking advantage of this detection network, we introduce the overall design of the proposed object detection framework, named "Deep Regionlets", as illustrated in Figure~\ref{fig::DeepRegionlet_overall}. 

The general architecture consists of an RSN and a deep regionlet learning module. 
In particular, the RSN is used to predict the transformation parameters to choose regions given a candidate bounding box, which is generated by the region proposal network. The regionlets are further learned within each selected region defined by the region selection network. The system is designed to be trained in a fully end-to-end manner using only the input images and ground truth bounding box. The RSN as well as the regionlet learning module can be simultaneously learned over each selected region given the detection window proposal.

\subsection{Region Selection Network \label{sec::region_selection}}

\begin{figure}[!bth]
\centering{
\includegraphics[width=0.8\columnwidth]{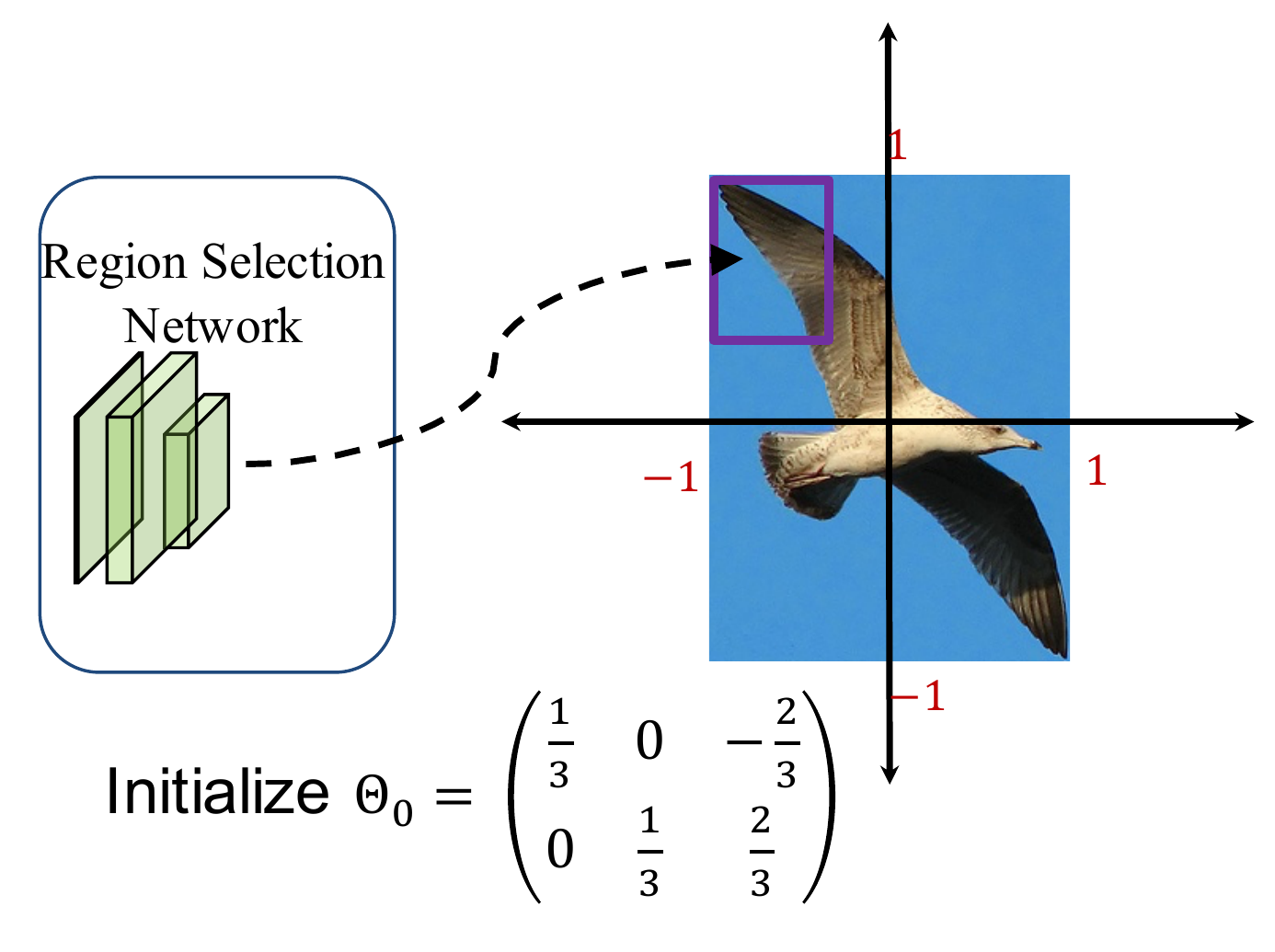}}
\caption{Example of initialization of one affine transformation parameter. Normalized affine transformation parameters $\Theta_0 = [\frac{1}{3}, 0, -\frac{2}{3}; 0, \frac{1}{3}, \frac{2}{3}]$ ($\theta_i \in [-1, 1]$) selects the top-left region in the $3\times 3$ evenly divided detection bounding box, shown as the purple rectangle.}
\label{fig::rsn_parameter_initialization}
\end{figure}

We design the RSN to have the following properties: 1) End-to-end trainable; 2) Simple structure; 3) Generate regions with arbitrary shapes.
Keeping these in mind, we design the RSN to predict a set of \emph{affine} transformation parameters. 
By using these affine transformation parameters, as well as not requiring the regions to be rectangular, we have more flexibility in modeling objects with arbitrary shapes and deformable parts.

Specifically, we design the RSN using a small neural network with three fully connected layers. The first two fully connected layers have output size of $256$, with ReLU activation. The last fully connected layer has the output size of six, which is used to predict the set of affine transformation parameters $\Theta = [ \theta_1, \theta_2, \theta_3; \theta_4, \theta_5, \theta_6 ]$.

Note that the candidate detection bounding boxes proposed by RSN have arbitrary sizes and aspect ratios. In order to address this difficulty, we use \emph{relative} positions and sizes of the selected region within a detection bounding box. The candidate bounding box generated by the RPN is defined by the top-left point ($w_0, h_0$), width $w$ and height $h$ of the box. We normalize the coordinates by the width $w$ and height $h$ of the box. As a result, we could use the normalized affine transformation parameters $\Theta = [ \theta_1, \theta_2, \theta_3; \theta_4, \theta_5, \theta_6 ]$ ($\theta_i \in [-1, 1]$) to evaluate one selected region within one candidate detection window at different sizes and aspect ratios without scaling images into multiple resolutions or using multiple-components to enumerate possible aspect ratios, like anchors~\cite{Ren_RHGS_TPAMI2016,Liu_LAESRFB_CVPR2016,Fu_FLRTB_2017}. 

\textbf{Initialization of Region Selection Network}: Taking advantage of \emph{relative} and \emph{normalized} coordinates, we initialize the RSN by equally dividing the whole detecting bounding box to several sub-regions, named as \emph{cell}s, without any overlap among them. Figure~\ref{fig::rsn_parameter_initialization} shows an example of initialization from one affine transformation
(\textit{i.e.} $3\times3$). The first cell, which is the top-left bin in the whole region (detection bounding box) could be defined by initializing the corresponding affine transformation parameter $\Theta_0 = [\frac{1}{3}, 0, -\frac{2}{3}; 0, \frac{1}{3}, \frac{2}{3}]$. The other eight of $3 \times 3$ cells are initialized in a similar way.

\subsection{Deep Regionlet Learning \label{sec::stn_regionlet}}

After regions are selected by the RSN, regionlets are further learned from the selected region defined by the normalized affine transformation parameters. Note that our motivation is to design the network to be trained in a fully end-to-end manner using only the input images and ground truth bounding boxes. Therefore, both the selected regions and regionlet learning should be able to be trained by CNN networks. Moreover, we would like the regionlets extracted from the selected regions to better represent objects with variable shapes and deformable parts.

Inspired by the spatial transform network~\cite{Jaderberg_JSZK_NIPS2015}, any parameterizable transformation including translation, scaling, rotation, affine or even projective transformation can be learned by a spatial transformer. In this section, we introduce our deep regionlet learning module to learn the regionlets in the selected region, which is defined by the affine transformation parameters.

More specifically, we aim to learn regionlets from one selected region defined by one affine transformation $\Theta$ to better match the shapes of objects. This is done with a selected region $R$ from RSN, transformation parameters $\Theta = [ \theta_1, \theta_2, \theta_3; \theta_4, \theta_5, \theta_6 ]$ and a set of feature maps $Z = \{Z_{i}, i = 1,\dots,n \}$. Without loss of generality, let $Z_{i}$ be one of the feature map out of the $n$ feature maps. A selected region $R$ is of size $w \times h$ with the top-left corner $(w_0, h_0)$. Inside the $Z_{i}$ feature maps, we propose the following regionlet learning module. 

Let $s$ denote the source and $t$ denote target, we define $(x_p^s, y_p^s)$ as the spatial location in original feature map $Z_{i}$ and $(x_p^s, y_p^s)$ as the spatial location in the output feature maps after spatial transformation. $U_{nm}^c$ is the value at location $(n, m)$ in channel $c$ of the input feature. The total output feature map $V$ is of size $H \times W$. Let $V(x_p^t, y_p^t, c| \Theta,R)$ be the output feature value at location ($x_p^t, y_p^t$) ($x_p^t\in [0, H]$, $y_p^t\in [0, W]$) in channel $c$, which is computed as

\small
\begin{equation}
\begin{split}
V(x_p^s, y_p^s, c| \Theta,R) = \sum_{n}^{H} & \sum_{m}^{M}  U_{nm}^c \max(0, 1 - | x_p^s - m|) \\ 
& \max(0, 1 - | y_p^s - n|) 
\label{eqn::stroi_bilinear}
\end{split}
\end{equation}
\normalsize

\textbf{Back Propagation through Spatial Transform}:
To allow back propagation of the loss through the regionlet learning module, we can define the gradients with respect to both feature maps and the region selection network. In this layer's \texttt{backward} function, we have partial derivative of the loss function with respect to both feature map variable $U_{mn}^c$ and affine transform parameter $\Theta = [ \theta_1, \theta_2, \theta_3; \theta_4, \theta_5, \theta_6 ]$. Motivated by~\cite{Jaderberg_JSZK_NIPS2015}, the partial derivative of the loss function with respect to the feature map is:

\small
\begin{equation}
\begin{split}
\frac{\partial V(x_p^s, y_p^s, c| \Theta,R)}{\partial U_{nm}^c} = \sum_{n}^{H} & \sum_{m}^{M} \max(0, 1 - | x_p^s - m|) \\
& \times \max(0, 1 - | y_p^s - n|)	
\end{split}
\label{eqn::stroi_partial_U}
\end{equation}
\normalsize

Moreover, during back propagation, we need to compute the gradient with respect to each affine transformation parameter $\Theta = [\theta_{1}, \theta_{2}, \theta_{3}; \theta_{4}, \theta_{5}, \theta_{6}]$. In this way, the region selection network could also be updated to adjust the selected region. We take $\theta_{1}$ as an example due to space limitations and similar derivative can be computed for other parameters $\theta_i (i = 2,\dots,6)$ respectively.

\small
\begin{equation}
\begin{split}
& \hspace{10mm} \frac{\partial V(x_p^s, y_p^s, c| \Theta,R)}{\partial \theta_{1}} = \frac{\partial V(x_p^s, y_p^s, c| \Theta,R)}{\partial x_{p}^s} \frac{\partial x_p^s}{\partial \theta_{1}} \\
& = x_p^t  \sum_{n}^{H} \sum_{m}^{M} U_{nm}^c \max(0, 1 - | y_p^s - n|) \times
\begin{cases}
0 \text{ if } |m - x_p^s| \ge 1 \\
1 \text{ if } m > x_p^s\\
-1 \text{ if } m < x_p^s
\end{cases}
\end{split}
\label{eqn::stroi_partial_theta}
\end{equation}
\normalsize

It is worth noting that $(x_p^t, y_p^t)$ are normalized coordinates in range $[-1, 1]$
so that it can to be scaled with respect to $w$ and $h$ with start position $(w_0, h_0)$. 

\begin{figure}[!tbh]
\centering{
\includegraphics[width=0.4\columnwidth]{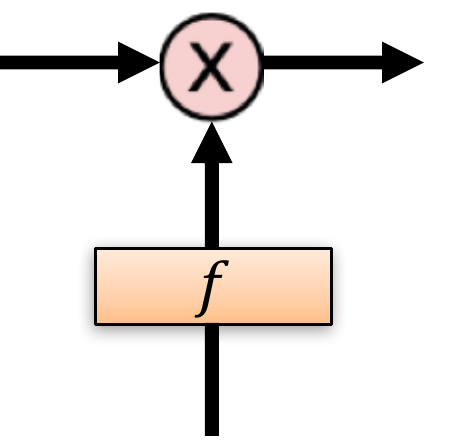}}
\caption{Design of the gating network. $f$ denotes the non-negative gate function (\textit{i.e.} sigmoid)}
\label{fig::gating_function}
\end{figure}

\textbf{Gating Network}: The gating network, which serves as a soft regionlet selector, is used to assgin regionlets with different weights and generate regionlet feature representation. We design a simple gating network using a fully connected layer with \texttt{sigmoid} activation, shown in Figure~\ref{fig::gating_function}. The output values of the gating network are within range of $[0, 1]$. Given the output feature maps $V(x_p^s, y_p^s, c| \Theta,R)$ described above, we use a fully connected layer to generate the same number of output as feature maps $V(x_p^s, y_p^s, c| \Theta,R)$, which is followed by an activation layer \texttt{sigmoid} to generate the corresponding weight respectively. The final feature representation is generated by the product of feature maps $V(x_p^s, y_p^s, c| \Theta,R)$ and their corresponding weights. 

\textbf{Regionlet Pool Construction}: Object deformations may occur at different scales. For instance, deformation could be caused by different body parts in person detection. Same number of regionlets (size $H\times W$) learned from small selected region have higher extraction density, which may lead to non-compact regionlet representation. In order to learn a \emph{compact}, \emph{efficient} regionlet representation, we further perform the pooling (\textit{i.e.} max/ave) operation over the feature maps $V(x_p^s, y_p^s, c| \Theta,R)$ of size ($H \times W$). We reap two benefits from the pool construction: (1) Regionlet representation is compact (small size). (2) Regionlets learned from different size of selected regions are able to represent such regions in the same efficient way, thus to handle object deformations at different scales.

\subsection{Relations to Recent Works\label{sec::related_discussion}}

Our deep regionlet approach is related to some recent works in different aspects. In this section, we discuss both similarities and differences in detail.

\textbf{Spatial Transform Networks (STN)} Jaderberg \textit{et al}.~\cite{Jaderberg_JSZK_NIPS2015} first proposed the spatial transformer module to provide spatial transformation capabilities into a deep neural network. It only learns \emph{one global parametric transformation} (scaling, rotations as well as affine transformation). Such learning is known to be difficult to apply on semi-dense vision tasks (\textit{e.g.}, object detection) and the transformation is on the entire feature map, which means the transformation is applied identically across all the regions in the feature map.

The proposed RSN learns a set of affine transformation and each transformation can be considered as the localization network in~\cite{Jaderberg_JSZK_NIPS2015}.
However, our regionlet learning is different from the image sampling~\cite{Jaderberg_JSZK_NIPS2015} method as it adopts a region-based parameter transformation and feature wrapping. By learning the transformation locally in the detection bounding box, our method provides the flexibility of learning a compact, efficient feature representation of objects with variable shape and deformable parts. 

\textbf{Deformable Part Model (DPM)~\textnormal{\cite{Felzenszwalb_FGM_CVPR2010}} and its deep learning extensions~\textnormal{\cite{Mordan_MTCH_2017,Dai_DQXLZHW_ICCV2017}}}
A Deformable Part Model~\cite{Felzenszwalb_FGM_CVPR2010} explicitly models the spatial deformations of object parts via latent variables. A root filter is learned to model the global appearance of the objects, while the part filters are designed to describe the local parts in the objects. However, DPM is a shallow model and the training process involves heuristic choices to select components and part sizes, making end-to-end training inefficient.

Both works~\cite{Dai_DQXLZHW_ICCV2017,Mordan_MTCH_2017} extend the DPM with end-to-end training in deep CNNs. Motivated by DPM~\cite{Felzenszwalb_FGMR_TPAMI2010} to allow parts to slightly move around their reference position (partition of the initial regions), they share the similar idea of learning part offsets\footnote{\cite{Dai_DQXLZHW_ICCV2017} uses term offset while~\cite{Mordan_MTCH_2017} uses term displacement} 
to model the local element and pool the features at their corresponding locations after the shift. While~\cite{Dai_DQXLZHW_ICCV2017,Mordan_MTCH_2017} show promising improvement over other deep learning-based object detectors~\cite{Girshick_G_ICCV2015,Ren_RHGS_TPAMI2016}, they still lack the flexibility of modeling non-rectangular objects with sharp shapes and deformable parts. 

It is noticeable that regionlet learning on the selected region is a generalization of Deformable RoI Pooling in~\cite{Dai_DQXLZHW_ICCV2017,Mordan_MTCH_2017}. First, we generalize the selected region to be non-rectangular by learning the affine transformation parameters. Such non-rectangular regions could provide the capabilities of \emph{scaling}, \emph{shifting} and \emph{rotation} around the original reference region. If we only enforce the region selection network to learn the shift, our regionlet learning mechanism would degenerate to similar deformable RoI pooling as in~\cite{Dai_DQXLZHW_ICCV2017,Mordan_MTCH_2017}

\textbf{Spatial-based RoI pooling}~\cite{Lazebnik_LSP_CVPR2006,Jia_JHD_2012CVPR,He_HZRS_ECCV20014}
Traditional spatial pyramid pooling~\cite{Lazebnik_LSP_CVPR2006} performs pooling over hand crafted regions at different scales. With the help of deep CNNs,~\cite{He_HZRS_ECCV20014} proposes to use spatial pyramid pooling in deep learning-based object detection. However, as the pooling regions over image pyramid still need to be carefully designed to learn the spatial layout of the pooling regions, therefore the end-to-end training is not well facilitated. 
In contrast, Our deep regionlet learning approach learns pooling regions end-to-end in deep CNNs. Moreover, the region selection step for learning regionlets accommodates different sizes of the regions. Hence, we are able to handle object deformations at different scales without generating the feature pyramid. 

\section{Experiments}

In this section, we present comprehensive experimental results of the proposed approach on two challenging benchmark datasets: PASCAL VOC~\cite{Everingham_EGWWZ_IJCV2010} and MS-COCO~\cite{Lin_LMBHPRDZ_ECCV2014}. 
There are in total $20$ categories of objects in PASCAL VOC~\cite{Everingham_EGWWZ_IJCV2010} dataset. We follow the common settings used in~\cite{Ren_RHGS_TPAMI2016,Bodla_BSCD_ICCV2017,Dai_DLHS_NIPS2016,Girshick_G_ICCV2015} to enable fair comparisons. 

More specifically, we train our deep model on (1) VOC $2007$ \texttt{trainval} and (2) union of VOC $2007$ \texttt{trainval} and $2012$ \texttt{trainval} and evaluate on VOC$2007$ \texttt{test}. We also report results on VOC $2012$ \texttt{test}, following the suggested settings in~\cite{Ren_RHGS_TPAMI2016,Bodla_BSCD_ICCV2017,Dai_DLHS_NIPS2016,Girshick_G_ICCV2015}. In addition, we report the results on the VOC$2007$ \texttt{test} split for ablation studies. 

MS-COCO~\cite{Lin_LMBHPRDZ_ECCV2014} contains $80$ object categories. Following the official settings in COCO website\footnote{\url{http://cocodataset.org/\#detections-challenge2017}}, , we use the COCO 2017 \texttt{trainval} split (union of $135$k images from \texttt{train} split and $5$k images from \texttt{val} split) for training. We report the COCO-style average precision (AP) on \texttt{test-dev} $2017$ split, which requires evaluation from the MS-COCO server.

For the base network, we choose both VGG-16~\cite{Simonyan_SZ_2014} and ResNet-101~\cite{He_HZRS_CVPR16} to demonstrate the generalization of our approach regardless of which network backbone we use. The \emph{\'{a} trous} algorithm~\cite{Long_LSD_CVPR2015,Mallat_M_1999} is adopted in stage $5$ of ResNet-101. Following the suggested settings in~\cite{Dai_DLHS_NIPS2016,Dai_DQXLZHW_ICCV2017}, we also set the pooling size to $7$ by changing the conv5 stage's effective stride from $32$ to $16$ to increase the feature map resolution. In addition, the first convolution layer with stride $2$ in the conv5 stage is modified to $1$. Both backbone networks are intialized with the pre-trained ImageNet~\cite{He_HZRS_CVPR16,Krizhevsky_KSH_2012NIPS} model. In the following sections, we report the results of a series of ablation experiments to understand the behavior of the proposed deep regionlet approach. Furthermore, we present comparisons with state-of-the-art detectors~\cite{Ren_RHGS_TPAMI2016,Dai_DLHS_NIPS2016,Dai_DQXLZHW_ICCV2017,He_HGDG_ICCV2017,Lin_LGGHD_ICCV2017,Lin_LDGHHB_CVPR2017} on both PASCAL VOC~\cite{Everingham_EGWWZ_IJCV2010} and MS COCO~\cite{Lin_LMBHPRDZ_ECCV2014} datasets.

\subsection{Ablation Study}\label{sec::ablation_study}

\begin{table*}[tbh]
\centering{
\setlength{\tabcolsep}{12pt}
\small
\resizebox{0.9\textwidth}{!}
{
\begin{tabular}{| c | c | c | c | c |  }
\hline
Methods & Global RSN  &  Offset-only RSN~\cite{Dai_DQXLZHW_ICCV2017,Mordan_MTCH_2017}  &  Non-gating & Ours \\
\hline
mAP@0.5($\%$) & 30.27 & 78.5 & 81.3 (+2.8) & 82.0 (+3.5)\\
\hline
\end{tabular}}
}
\caption{Ablation study of each component in deep regionlet approach. Output size $H\times W$ is set to $4\times 4$ for all the baselines}
\label{tab::each_component}
\end{table*}

\begin{table*}[tbh]
\centering{
\setlength{\tabcolsep}{12pt}
\small
\resizebox{0.9\textwidth}{!}{
\begin{tabular}[b]{|c|*{5}{c|}}\hline
\backslashbox{$\#$ of Regions}{Regionlets Density}
& $2\times 2$ & $3\times 3$ & $4\times 4$& $5 \times 5$& $6\times 6$ \\ 
\hline
$4(2\times 2)$ regions &  78.0  & 79.2  &   79.9  & 80.2    & 80.3 \\
\hline
$9 (3\times 3)$ regions&  79.6  & 80.3  &   80.9  &  81.5 & 81.3 \\
\hline
$16 (4\times 4)$ regions& 80.0  & 81.0  & 82.0  & 81.6  &   80.8 \\
\hline
\end{tabular}}
}
\caption{Results of ablation studies when a region selection network (RSN) selects different number of regions and regionlets are learned at different level of density.}
\label{tab::numOfRegions_regionletsDensity}
\end{table*}

For a fair comparison, we adopt ResNet-101 as the backbone network for ablation studies. We train our model on the union set of VOC $2007+2012$ \texttt{trainval} and evaluate on the VOC$2007$ \texttt{test} set. The shorter side of image is set to be $600$ pixels, as suggested in~\cite{Girshick_G_ICCV2015,Ren_RHGS_TPAMI2016,Dai_DLHS_NIPS2016}. The training is performed for $60$k iterations with an effective mini-batch size $4$ on $4$ GPUs, where the learning rate is set at $10^{-3}$ for the first $40$k iterations and at $10^{-4}$ for the remaining $20$k iterations. First we investigate the proposed approach to understand each component (1) RSN, (2) Deep regionlet learning and (3) Soft regionlet selection by comparing it with several baselines: 

\begin{enumerate}
  \item Global RSN. RSN only selects one global region and it is initialized as identity transformation (\textit{i.e.} $\Theta_0 = [1, 0, 0; 0, 1, 0]$). This is equivalent to global regionlet learning within the RoI.
  \item Offset-only RSN. We set the RSN to only learn the offset by enforcing $\theta_1, \theta_2, \theta_4, \theta_5$ not to change during the training process. In this way, the region selection network only selects the rectangular region with offsets to the initialized region. This baseline is similar to the Deformable RoI Pooling in ~\cite{Dai_DQXLZHW_ICCV2017} and ~\cite{Mordan_MTCH_2017}.
  \item Non-gating selection: Deep regionlet without soft selection. No soft regionlet selection is performed after the regionlet learning. In this case, each regionlet learned has the same contribution to the final feature representation.
\end{enumerate}

Results are shown in Table~\ref{tab::each_component}. First, when the region selection network only selects one global region, the RSN reduces to the single localization network~\cite{Jaderberg_JSZK_NIPS2015}. In this case, regionlets will be extracted in a global manner. It is interesting to note that selecting only one region by the region selection network is able to converge, which is different from~\cite{Ren_RHGS_TPAMI2016,Dai_DLHS_NIPS2016}. However, the performance is extremely poor. This is because no discriminative regionlets could be explicitly learned within the region. More importantly, when we compare our approach and offset-only RSN with global RSN, the results clearly demonstrate that the RSN is \emph{indispensable} in the deep regionlet approach. 

Moreover, offset-only RSN could be viewed as similar to deformable RoI pooling in~\cite{Dai_DQXLZHW_ICCV2017,Mordan_MTCH_2017}. These methods all learn the offset of the rectangle region with respect to its reference position, which lead to improvement over~\cite{Ren_RHGS_TPAMI2016}. However, non-gating selection outperforms offset-only RSN by $2.8\%$ while selecting the non-rectangular region. The improvement demonstrates that non-rectangular region selection could provide more flexibility around the original reference region, thus could better model the non-rectangular objects with sharp shapes and deformable parts. Last but not least, by using the gate function to perform soft regionlet selection, the performance can be further improved by $0.7\%$.

Next, we present ablation studies on the following questions in order to understand more deeply on the region selection network and regionlet learning module: 
\begin{enumerate}
\item How many regions should we learn using the region selection network?
\item How many regionlets should we learn in a selected region (density is of size $H\times W$)? 
\end{enumerate}

\noindent\textbf{How many regions should we learn using the region selection network?} We investigate how the detection performance varies when different number of regions are selected by the region selection network. All the regions are initialized as described in Section~\ref{sec::region_selection} without any overlap between regions. Without loss of generality, we report results for $4 ( 2\times 2)$, $9 (3 \times 3)$ and $16 (4 \times 4)$ regions in Table~\ref{tab::numOfRegions_regionletsDensity}. We observe that the mean AP increases when the number of selected regions is increased from $4(2 \times 2)$ to $9 (3 \times 3)$ for a fixed regionlets learning number, but gets saturated with $16(4 \times 4)$ selected regions. 

\noindent \textbf{How many regionlets should we learn in one selected region?} Next, we investigate how the detection performance varies when different number of regionlets are learned in one selected region by varying $H$ and $W$. Without loss of generality, we set $H = W$ and vary the $H$ value from $2$ to $6$. In Table~\ref{tab::numOfRegions_regionletsDensity}, we report results when we set the number of regionlets at $4 (2 \times 2)$, $9 (3 \times 3)$, $16 (4 \times 4)$, $25 (5 \times 5)$, $36 (6 \times 6)$ before the regionlet pooling construction. 

First, it is observed that increasing the number of regionlets from $4(2\times 2)$ to $25 (5\times 5)$ results in improved performance. As more regionlets are learned from one region, more spatial and shape information from objects could be learned. The proposed approach could achieve the best performance when regionlets are extracted at $16 (4\times 4)$ or $25 (5 \times 5)$ density level. It is also interesting to note that when the density increases from $25(5 \times 5)$ to $36 (6 \times 6)$, the performance degrades slightly. When the regionlets are learned at a very high density level, some redundant spatial information may be learned without being useful for detection, thus affecting the region proposal-based decision to be made. In all the experiments, we present the results from $16$ selected regions from the RSN and set output size $H \times W = 4 \times 4$. 


\subsection{Experiments on PASCAL VOC}

\begin{table*}[t]
\centering{
\setlength{\tabcolsep}{13pt}
\small
\resizebox{0.95\textwidth}{!}
{
\begin{tabular}{| c | c | c | c | c |  }
\hline
Methods &training data & mAP@0.5($\%$) &  training data & mAP@0.5($\%$) \\
\hline
Regionlet~\cite{Wang_WYZL_ICCV2013} & 07 &  41.7 &  07 + 12 & N/A \\
\hline 
Faster R-CNN~\cite{Ren_RHGS_TPAMI2016} & 07 & 70.0 & 07 + 12 & 73.2 \\
\hline
R-FCN~\cite{Dai_DLHS_NIPS2016} & 07 & 69.6  & 07 + 12 & 76.6 \\
\hline
SSD 512~\cite{Liu_LAESRFB_CVPR2016} & 07 & 71.6 &  07 + 12 & 76.8 \\
\hline
Soft-NMS~\cite{Bodla_BSCD_ICCV2017} & 07 & 71.1 & 07 + 12 & 76.8 \\
\hline
Ours & 07 & \textbf{\emph{73.0}} & 07 + 12 & \textbf{\emph{79.2}}\\
\hline
Ours$^\S$ & 07 & \textbf{73.8} & 07 + 12 & \textbf{80.1} \\
\hline
\end{tabular}}
}
\caption{Detection results on PASCAL VOC using VGG16 as backbone architecture. Training data: "07": VOC$2007$ \texttt{trainval}, "07 + 12": VOC $2007$ and $2012$ \texttt{trainval}. Ours$^\S$ denotes applying the soft-NMS~\cite{Bodla_BSCD_ICCV2017} in the test stage.}
\label{tab::VOC2007_VGG16}
\end{table*}

\begin{table*}[t]
\centering{
\setlength{\tabcolsep}{4pt}
\small
\resizebox{0.95\textwidth}{!}
{
\begin{tabular}{| c | c | c | c |}
\hline
Methods & mAP@0.5 / @0.7($\%$) & Methods & mAP@0.5 / @0.7($\%$) \\
\hline
Faster R-CNN~\cite{Ren_RHGS_TPAMI2016} &  78.1 / 62.1 & SSD~\cite{Liu_LAESRFB_CVPR2016} &  76.8 / N/A  \\
\hline 
DP-FCN~\cite{Mordan_MTCH_2017} &  78.1 / N/A & ION~\cite{Bell_BZBG_CVPR2016} & 79.4 / N/A \\
\hline
LocNet~\cite{Gidaris_GK_CVPR2016} & 78.4 / N/A & Deformable ConvNet~\cite{Dai_DQXLZHW_ICCV2017} &  78.6 / 63.3 \\
\hline
Deformable ROI Pooling~\cite{Dai_DQXLZHW_ICCV2017} & 78.3 / 66.6 & D-F-RCNN~\cite{Dai_DQXLZHW_ICCV2017} & 79.3 / 66.9 \\
\hline
Ours & \textbf{\textit{82.0}} / \textbf{\textit{67.0}} & Ours$^\S$ & \textbf{83.1} / \textbf{67.9} \\
\hline
\end{tabular}}
}
\caption{Detection results on PASCAL VOC using ResNet-101~\cite{He_HZRS_CVPR16} as backbone acchitecture. Training data: union set of VOC $2007$ and $2012$ \texttt{trainval}. Ours$^\S$ denotes applying the soft-NMS~\cite{Bodla_BSCD_ICCV2017} in the test stage.}
\label{tab::VOC2007_ResNet101}
\end{table*}

\begin{table*}[bht]
\centering{
\setlength{\tabcolsep}{10pt}
\small
\resizebox{0.95\textwidth}{!}
{
\begin{tabular}{| c | c | c | c | c | c |}
\hline
Methods & FRCN~\cite{Ren_RHGS_TPAMI2016} &  YOLO9000~\cite{Redmon_RF_CVPR2017}  &  FRCN OHEM & DSSD~\cite{Fu_FLRTB_2017} &  SSD$^\ast$~\cite{Liu_LAESRFB_CVPR2016} \\
\hline
mAP@0.5($\%$) & 73.8 & 73.4 & 76.3 & 76.3 & 78.5\\
\hline
Methods & ION~\cite{Bell_BZBG_CVPR2016} & R-FCN~\cite{Dai_DLHS_NIPS2016} & DP-FCN~\cite{Mordan_MTCH_2017} &  Ours & Ours$^\S$ \\
\hline 
mAP@0.5($\%$)  & 76.4 & 77.6 & 79.5 & \textbf{\emph{80.4}} & \textbf{81.2} \\
\hline
\end{tabular}}
}
\caption{Detection results on VOC$2012$ \texttt{test} set using training data "07++12": 2007 \texttt{trainvaltest} and 2012 \texttt{trainval}. SSD$^\ast$ denotes the new data augmentation. Ours$^\S$ denotes applying the soft-NMS~\cite{Bodla_BSCD_ICCV2017} in the test stage.}
\label{tab::VOC2012}
\end{table*}

\begin{table*}[t]
\centering{
\setlength{\tabcolsep}{4pt}
\resizebox{0.95\textwidth}{!}
{
\begin{tabular}{| c | c | c | c | c | c | c |}
\hline
Methods & Training Data & mmAP $0.5$:$0.95$ & mAP @0.5 & mAP small & mAP medium & mAP large \\
\hline
Faster R-CNN~\cite{Ren_RHGS_TPAMI2016} &  trainval & 24.4 &  45.7 & 7.9 & 26.6 & 37.2 \\
\hline
SSD$^\ast$\cite{Liu_LAESRFB_CVPR2016} & trainval & 31.2 &  50.4 & 10.2 & 34.5 & 49.8 \\ 
\hline  
DSSD~\cite{Fu_FLRTB_2017} & trainval & 33.2 & 53.5 & 13.0 & 35.4 & 51.1 \\
\hline 
R-FCN~\cite{Dai_DLHS_NIPS2016} & trainval & 30.8 & 52.6 & 11.8 & 33.9 & 44.8 \\
\hline 
D-F-RCNN~\cite{Dai_DQXLZHW_ICCV2017} & trainval & 33.1 & 50.3 & 11.6 & 34.9 & 51.2 \\
\hline
D-R-FCN~\cite{Dai_DQXLZHW_ICCV2017} & trainval & 34.5 & 55.0 & 14.0 & 37.7 & 50.3 \\
\hline
Mask R-CNN~\cite{He_HGDG_ICCV2017} & trainval & 38.2 & 60.3 & 20.1 & 41.1 & 50.2  \\
\hline 
RetinaNet500~\cite{Lin_LGGHD_ICCV2017} & trainval &  34.4 & 53.1 & 14.7 & 38.5 & 49.1 \\ 
\hline 
Ours & trainval & \textbf{39.3} & 59.8 & \textbf{21.7} & \textbf{43.7} &  50.9 \\
\hline
\end{tabular}}
}
\caption{Object detection results on MS COCO $2017$ \texttt{test-dev} using ResNet-101 backbone. Training data: $2017$ \texttt{train} and \texttt{val} set. SSD$^\ast$ denotes the new data augmentation.}
\label{tab::COCO_ResNet101}
\end{table*}

In this section, we compare our results with a traditional regionlet method~\cite{Wang_WYZL_ICCV2013} and several state-of-the-art deep learning-based object detectors as follows: Faster R-CNN~\cite{Ren_RHGS_TPAMI2016}, SSD~\cite{Liu_LAESRFB_CVPR2016}, R-FCN~\cite{Dai_DLHS_NIPS2016}, soft-NMS~\cite{Bodla_BSCD_ICCV2017}, DP-FCN~\cite{Mordan_MTCH_2017} and D-F-RCNN/D-R-FCN~\cite{Dai_DQXLZHW_ICCV2017}. 

We follow the standard settings as in~\cite{Ren_RHGS_TPAMI2016,Dai_DLHS_NIPS2016,Bodla_BSCD_ICCV2017,Dai_DQXLZHW_ICCV2017}  and report mean average precision (mAP) scores using IoU thresholds at $0.5$ and $0.7$. 
For the first experiment, while training from VOC $2007$ \texttt{trainval}, we use a learning rate of $10^{-3}$ for the first $40$k iterations, then decrease it to $10^{-4}$ for the remaining $20$k iterations with a single GPU. Next, due to more training data, an increase in the number of iterations is needed on the union of VOC $2007$ and VOC $2012$ \texttt{trainval}. We perform the same training process as described in Section~\ref{sec::ablation_study}. Moreover, we use $300$ RoIs at test stage from a single-scale image testing and set the shorter side of the image to be $600$.
For a fair comparison, we do not deploy the multi-scale training/testing or online hard example mining(OHEM)~\cite{Shrivastava_SGG_CVPR2016}, although it is shown in~\cite{Bodla_BSCD_ICCV2017,Dai_DQXLZHW_ICCV2017} that such enhancements could enhance the performance. 

The results on VOC$2007$ \texttt{test} using VGG$16$~\cite{Simonyan_SZ_2014} backbone are shown in Table~\ref{tab::VOC2007_VGG16}. We first compare with a traditional regionlet method~\cite{Wang_WYZL_ICCV2013} and several state-of-the-art object detectors~\cite{Ren_RHGS_TPAMI2016,Liu_LAESRFB_CVPR2016,Bodla_BSCD_ICCV2017} when training using small size dataset (VOC $2007$ \texttt{trainval}). Next, we evaluate our method as we increase the training dataset (union set of VOC $2007$ and $2012$ \texttt{trainval}). With the power of deep CNNs, the deep regionlet approach significantly improves the detection performance over the traditional regionlet method~\cite{Wang_WYZL_ICCV2013}. We also observe that more data always helps. Moreover, it is encouraging that soft-NMS~\cite{Bodla_BSCD_ICCV2017} is only applied in the test stage without modification in the training stage, which could directly improve over~\cite{Ren_RHGS_TPAMI2016} by $1.1\%$. In summary, our method consistently outperform all the compared methods and the performance could be further improved if we replace NMS with soft-NMS~\cite{Bodla_BSCD_ICCV2017}

Next, we change the network backbone from VGG16~\cite{Simonyan_SZ_2014} to ResNet-$101$~\cite{He_HZRS_CVPR16} and present corresponding results in Table~\ref{tab::VOC2007_ResNet101}. In addition, we also compare with D-F-RCNN/D-R-FCN~\cite{Dai_DQXLZHW_ICCV2017} and DP-FCN~\cite{Mordan_MTCH_2017}.

First, compared to the performance in Table~\ref{tab::VOC2007_VGG16} using VGG16~\cite{Simonyan_SZ_2014} network, the mAP can be significantly increased by using deeper networks like ResNet-101~\cite{He_HZRS_CVPR16}. Second, comparing with DP-FCN~\cite{Mordan_MTCH_2017} and Deformable ROI Pooling in~\cite{Dai_DQXLZHW_ICCV2017}\footnote{~\cite{Dai_DQXLZHW_ICCV2017} reported best result using OHEM, We only compare the results reported in~\cite{Dai_DQXLZHW_ICCV2017} without deploying OHEM.}, we outperform these two methods by $\mathbf{3.9\%}$ and $\mathbf{2.7\%}$ respectively. This provides the empirical support that our deep regionlet learning method could be treated as a \emph{generalization} of Deformable RoI Pooling in~\cite{Dai_DQXLZHW_ICCV2017,Mordan_MTCH_2017}, as discussed in Section~\ref{sec::related_discussion}. In addition, the results demonstrate that selecting \emph{non-rectangular} regions from our method provides more capabilities including \emph{scaling}, \emph{shifting} and \emph{rotation} to learn the feature representations. In summary, our method achieves state-of-the-art performance on the object detection task when using ResNet-$101$ as backbone network. 

Results evaluated on VOC $2012$ \texttt{test} are shown in Table~\ref{tab::VOC2012}. We follow the same settings as in~\cite{Dai_DLHS_NIPS2016,Ren_RHGS_TPAMI2016,Fu_FLRTB_2017,Liu_LAESRFB_CVPR2016,Mordan_MTCH_2017} and train our model using VOC"07++12": VOC $2007$ \texttt{trainvaltest} and $2012$ \texttt{trainval} set. It can be seen that our method outperform all the competing methods. In particular, we outperform DP-FCN~\cite{Mordan_MTCH_2017}, which further proves the generalization of our method over~\cite{Mordan_MTCH_2017}.

\subsection{Experiments on MS COCO}

In this section, we evaluate the proposed deep regionlet approach on the MS COCO~\cite{Lin_LMBHPRDZ_ECCV2014} dataset and compare with other state-of-the-art object detectors: Faster R-CNN~\cite{Ren_RHGS_TPAMI2016}, SSD~\cite{Liu_LAESRFB_CVPR2016}, R-FCN~\cite{Dai_DLHS_NIPS2016}, D-F-RCNN/D-R-FCN~\cite{Dai_DQXLZHW_ICCV2017}, Mask R-CNN~\cite{He_HGDG_ICCV2017}, RetinaNet~\cite{Lin_LGGHD_ICCV2017}. 

We adopt ResNet-101 as the backbone architecture of all the methods for a fair comparison. Following the settings in~\cite{He_HGDG_ICCV2017,Dai_DQXLZHW_ICCV2017,Lin_LGGHD_ICCV2017,Dai_DLHS_NIPS2016}, we set the shorter edge of the image to $800$ pixels. Training is performed for $280$k iterations with an effective mini-batch size $8$ on 8 GPUs. We first train the model with a learning rate of $10^{-3}$ for the first $160$k iterations, followed by learning rates of $10^{-4}$ and $10^{-5}$ subsequent for another $80$k iterations and the last $40$k iterations respectively. Five scales and three aspect ratios are deployed as anchors. We report results using either the released models or the code from the original authors. It is noted that we only deploy single-scale image training without the iterative bounding box average, although these enhancements could further boost performance (mmAP). 

Table~\ref{tab::COCO_ResNet101} shows the results on 2017 \texttt{test-dev} set, which contains $20,288$ images. Compared with the baseline methods Faster R-CNN~\cite{Ren_RHGS_TPAMI2016}, R-FCN~\cite{Dai_DLHS_NIPS2016} and SSD~\cite{Liu_LAESRFB_CVPR2016}, both D-F-RCNN/D-R-FCN~\cite{Dai_DQXLZHW_ICCV2017} and our method provides significant improvements over~\cite{Ren_RHGS_TPAMI2016,Dai_DLHS_NIPS2016,Liu_LAESRFB_CVPR2016} (+$3.7\%$ and +$8.5\%$). Moreover, it can be seen that the proposed method outperforms D-F-RCNN/D-R-FCN~\cite{Dai_DQXLZHW_ICCV2017} by a wide margin($\sim$$\mathbf{4\%}$). This observation further supports that our deep regionlet learning module could be treated as a \emph{generalization} of Deformable RoI Pooling in~\cite{Dai_DQXLZHW_ICCV2017,Mordan_MTCH_2017}. It is also noted that although most recent state-of-the-art object detectors such as Mask R-CNN~\cite{He_HGDG_ICCV2017} utilize multi-task training with segmentation labels, we still outperform Mask R-CNN~\cite{He_HGDG_ICCV2017} by $1.1\%$.  
In addition, the focal loss in~\cite{Lin_LGGHD_ICCV2017}, which overcomes the obstacle caused by the imbalance of positive/nagetive samples, is complimentary to our method. We believe it can be integrated into our method to further boost performance. In summary, compared with Mask R-CNN~\cite{He_HGDG_ICCV2017} and RetinaNet\footnote{\cite{Lin_LGGHD_ICCV2017} reported best result using multi-scale training for 1.5$\times$ longer iterations, we only compare the results without scale jitter during training. In addition, we only compare the results in~\cite{He_HGDG_ICCV2017} using ResNet-101 backbone for fair comparison.}~\cite{Lin_LGGHD_ICCV2017}, our method achieves competitive performance over state-of-the-art on MS COCO when using ResNet-$101$ as a backbone network. 

\section{Conclusion}

In this paper, we present a novel deep regionlet-based approach for object detection. The proposed RSN can select \emph{non-rectangular} regions within the detection bounding box, and hence an object with rigid shape and deformable parts can be better modeled. We also design the deep regionlet learning module so that both the selected regions and the regionlets can be learned simultaneously. Moreover, the proposed system can be trained in a fully end-to-end manner without additional efforts. Finally, we extensively evaluate our approach on two detection benchmarks and experimental results show competitive performance over state-of-the-art.

\section{Acknowledgement}

This research is based upon work supported by the Intelligence Advanced Research Projects Activity (IARPA) via Department of Interior/Interior Business Center (DOI/IBC) contract number D17PC00345. The U.S. Government is authorized to reproduce and distribute reprints for Governmental purposes not withstanding any copyright annotation theron. Disclaimer: The views and conclusions contained herein are those of the authors and should not be interpreted as necessarily representing the official policies or endorsements, either expressed or implied of IARPA, DOI/IBC or the U.S. Government.​

We thank the reviewers for their valuable comments and suggestions.

{
\small
\bibliographystyle{ieee}
\bibliography{cvpr2018}
}

\end{document}